\documentclass[conference]{IEEEtran}
\IEEEoverridecommandlockouts
\usepackage{cite}
\usepackage{amsmath,amssymb,amsfonts}
\usepackage{algorithmic}
\usepackage{graphicx}
\usepackage{array}
\usepackage{float}
\usepackage[edges]{forest}
\usepackage[numbers]{natbib}
\usepackage{enumerate}
\usepackage{dirtree}
\usepackage{textcomp}
\usepackage{xcolor}
\def\BibTeX{{\rm B\kern-.05em{\sc i\kern-.025em b}\kern-.08em
    T\kern-.1667em\lower.7ex\hbox{E}\kern-.125emX}}
\begin{document}

\title{Root Cause Analysis Method Based on Large Language Models with Residual Connection Structures\\

\thanks{https://www.aiops.cn/gitlab/aiops-live-benchmark/aiopschallenge2025}
}

\author{\IEEEauthorblockN{1\textsuperscript{st} Liming Zhou}
\IEEEauthorblockA{\textit{dept. name of organization (of Aff.)} \\
\textit{name of organization (of Aff.)}\\
Beijing, China \\
email address or ORCID}
\and
\IEEEauthorblockN{2\textsuperscript{nd} Ailing Liu}
\IEEEauthorblockA{\textit{dept. name of organization (of Aff.)} \\
\textit{name of organization (of Aff.)}\\
Beijing, China \\
email address or ORCID}
\and
\IEEEauthorblockN{3\textsuperscript{rd} Hongwei Liu}
\IEEEauthorblockA{\textit{dept. name of organization (of Aff.)} \\
\textit{name of organization (of Aff.)}\\
Beijing, China \\
email address or ORCID}
\and
\IEEEauthorblockN{4\textsuperscript{th} Min He*}
\IEEEauthorblockA{\textit{dept. name of organization (of Aff.)} \\
\textit{name of organization (of Aff.)}\\
Beijing, China \\
email address or ORCID}
\and
\IEEEauthorblockN{5\textsuperscript{th} Heng Zhang}
\IEEEauthorblockA{\textit{dept. name of organization (of Aff.)} \\
\textit{name of organization (of Aff.)}\\
City, Country \\
email address or ORCID}
\and
\IEEEauthorblockN{6\textsuperscript{th} Given Name Surname}
\IEEEauthorblockA{\textit{dept. name of organization (of Aff.)} \\
\textit{name of organization (of Aff.)}\\
City, Country \\
email address or ORCID}
}

\maketitle

\begin{abstract}
Root cause localization remain challenging in complex and large-scale microservice architectures. The complex fault propagation among microservices and the high dimensionality of telemetry data, including metrics, logs, and traces, limit the effectiveness of existing root cause analysis (RCA) methods. In this paper, a residual-connection-based RCA method using large language model (LLM), named RC-LLM, is proposed. A residual-like hierarchical fusion structure is designed to integrate multi-source telemetry data, while the contextual reasoning capability of large language models is leveraged to model temporal and cross-microservice causal dependencies. Experimental results on CCF-AIOps microservice datasets demonstrate that RC-LLM achieves strong accuracy and efficiency in root cause analysis.
\end{abstract}


\section{Introduction}
With the widespread adoption of cloud computing and DevOps practices, modern enterprise software systems are increasingly transformed into microservice architectures \cite{balalaie2016microservices}. In such architectures, monolithic applications are decomposed into loosely coupled and independently deployable services, which significantly improves development agility, scalability, and resource utilization. However, the distributed nature of microservices introduces substantial operational complexity. In large-scale systems composed of hundreds of services, failures are no longer isolated within individual services but can propagate rapidly through complex call chains and message queues, resulting in intricate fault propagation paths.

When performance degradation or system failures occur, rapid and accurate root cause analysis (RCA) becomes a critical challenge for operators. Traditional RCA approaches based on manual expertise, log inspection, or predefined rules are inadequate for microservice environments characterized by high concurrency, large-scale heterogeneous telemetry data, including metrics, logs, and traces, and dynamically evolving dependencies. Delayed fault localization often leads to prolonged service outages, increased mean time to repair (MTTR), and significant business impact. Therefore, the development of an intelligent RCA approach capable of automatically and efficiently identifying the root cause from noisy data is essential for ensuring the reliability of microservice systems.

\begin{figure}
    \centering
    \includegraphics[width=1\linewidth]{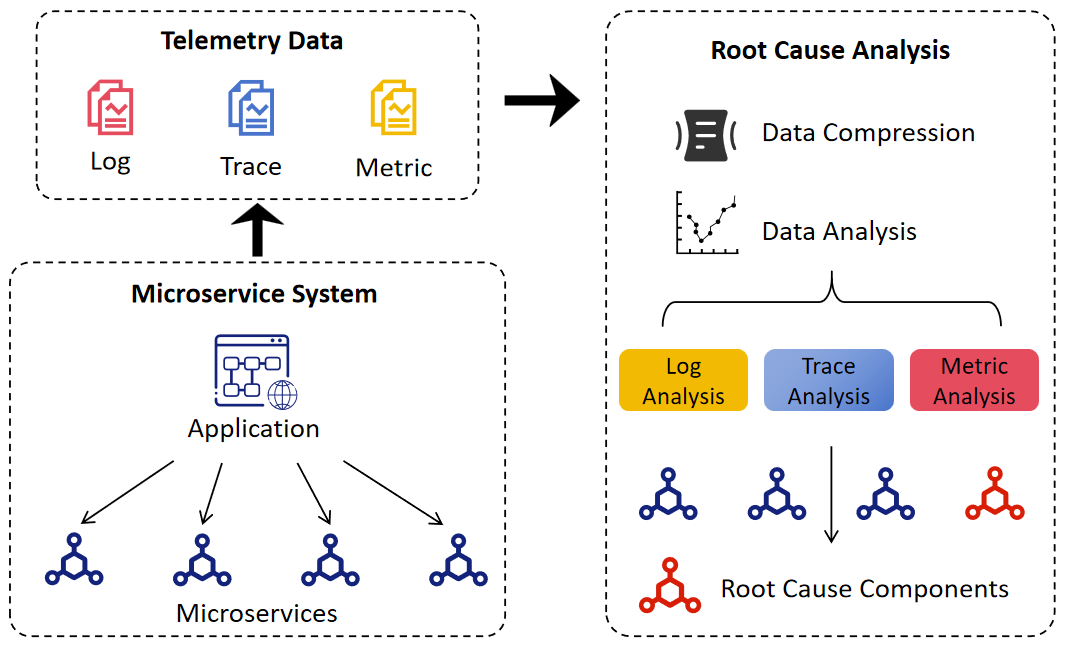}
    \caption{Root Cause Analysis for Microservice-based Systems.}
    \label{fig1}
    \vspace{-0.5em}
\end{figure}

To address the above challenges, a residual-connection-based RCA method using LLM, named RC-LLM, is proposed. In this method, the RCA task is formulated as a deep temporal causal reasoning problem and is addressed by leveraging the strong sequence modeling capability of LLMs. To effectively integrate heterogeneous telemetry data and improve information flow, a multi-source integration strategy with residual connection structures is designed.

The main contributions of this paper are summarized as follows:
\begin{enumerate}[(1)]
    \item We proposed A residual-connection-based RCA method for microservice architectures, named RC-LLM. Multi-source telemetry data are integrated to address the limitations of existing approaches in modeling complex fault propagation.
    \item  A multi-source integration strategy with residual connection structures is designed for organizing traces, metrics, and logs as structured inputs for LLM reasoning. This design promotes effective information preservation and reuse across data sources, facilitating the modeling of long-range and cross-service dependencies in microservice systems.
    \item  Extensive experiments conducted on a realistic microservice system data demonstrate that RC-LLM achieves strong accuracy and efficiency in root cause analysis.
\end{enumerate}

\section{Related Work}
Root cause analysis in microservice architectures, similar to the evolution of other computer science tasks, can be broadly categorized into three stages: (A) rule-based and expert-driven methods; (B) machine learning-based methods; and (C) LLM-based approaches that leverage powerful sequential feature encoding and contextual reasoning capabilities to address complex challenges in AIOps \cite{zhang2025survey}.

\subsection{Rule-Based and Expert-Driven RCA Methods}
Early RCA approaches primarily relied on operational expertise, predefined rules, and structured data analysis. These methods include threshold-based approaches, in which anomalies are detected using static rules or thresholds and root causes are inferred by matching predefined failure signatures, such as alerts triggered by CPU utilization, latency, or error rate metrics. \citet{nguyen2011pal} proposed a non-intrusive anomaly localization system that detects critical change points in system-level metrics and sorts them temporally to infer anomaly propagation patterns. \citet{marvasti2013anomaly} introduced a data-driven anomaly event correlation approach that models temporal co-occurrence relationships among historical anomaly events to enable root cause and extreme event identification. However, in microservice environments with numerous services and continuously evolving dependencies, maintaining a complete and accurate rule set becomes impractical, and static rules exhibit limited adaptability to dynamic system behaviors.

Dependency-graph-based approaches have also been extensively studied. In these methods, service dependency graphs  or causality graphs are constructed based on inter-service call relationships. \citet{brandon2020graph} built a directed dependency graph and performed anomaly propagation analysis to localize root causes in microservice systems. \citet{xin2023causalrca} employed gradient-based causal structure learning to generate weighted causal graphs and applied a PageRank-based inference strategy to identify faulty components. \citet{wu2021microdiag} inferred fine-grained root causes by modeling anomaly propagation among metrics using dependency and causality graphs. \citet{aubet2018graph} proposed a graph-based anomaly detection system for distributed IoT microservices that adaptively updates the communication model to capture temporal changes in deployments. After anomalies are detected, fault localization is typically conducted through graph traversal or causal inference on the constructed graphs. Despite their effectiveness, these approaches heavily rely on the completeness and accuracy of dependency graphs and often fail to capture implicit fault propagation caused by resource contention or shared component failures beyond explicit call relationships.

\subsection{Machine Learning-Based RCA Methods}
In recent years, with the rapid growth of telemetry data (metrics, logs, and traces) and advances in computing capabilities, machine learning techniques have been widely adopted for automated RCA to overcome the limitations of traditional approaches. Many studies focus on metric-based or log-based anomaly methods, in which anomaly detection and correlation analysis are performed on the corresponding  data sources. \citet{wang2021groot} proposed an event-graph-based RCA method that constructs real-time causality graphs using heterogeneous events derived from metrics and logs as nodes, and infers causal relationships in industrial microservice systems by incorporating domain-specific rules and heuristics. Additionally, trace-based methods for RCA have been increasingly explored. \citet{liu2020unsupervised} proposed an unsupervised trace anomaly detection system for microservices, in which service-level vectors are constructed to encode execution paths and response times, and Bayesian networks are employed to learn normal patterns. \citet{meng2021detecting} introduced an anomaly detection approach that constructs a baseline of normal traces and measures trace similarity using tree edit distance to detect structural anomalies, while principal component analysis is applied to identify anomalous components with significant delay variations. 

Neural network-based methods such as temporal models (e.g., LSTM \cite{hochreiter1997long}, GRU \cite{chung2014empirical}), and autoencoders (e.g., Transformer \cite{vaswani2017attention}, BERT \cite{devlin2019bert}), have also been widely adopted for RCA. \citet{yang2022micromilts} proposed a root cause localization method based on mutual information and LSTM, in which performance metric deviations and real-time call differences are combined to construct service dependency graphs, and fault localization is performed using a PageRank-based inference strategy. \citet{zhang2022fault} proposed a fault localization method that integrates system logs and monitoring metrics. This method utilizes Transformer models to capture sequence features from the merged log and monitoring metric data to detect anomalies in microservices. Despite their effectiveness, these methods often overlook the structured and relational information inherent in log and trace data.

In addition, graph neural networks (GNNs) \cite{scarselli2008graph} have been widely explored for microservice RCA due to their ability to model graph-structured data. \citet{li2022actionable} constructed a Failure Dependency Graph (FDG) with candidate failure units (groups of indicative metrics on components) as nodes and apply Graph Attention Networks (GAT) to capture dependencies and unit-level features on the graph, thereby achieving localization of recurring failures. The Sleuth \cite{gan2023sleuth} employs GNNs to model the causal influences across temporal intervals in trace data, and clusters traces using a distance-based metric, thereby effectively reducing the number of traces needed for RCA. \citet{yen2022graph} combined GNNs with graph structure learning (GSL), leveraging GSL models to infer latent dependencies from data with a large numbers of nodes and generate information-rich graph representations for GNNs to identify root causes. Nevertheless, existing GNN-based methods still face challenges in handling the dynamic nature of microservice architectures and heterogeneous data integration across metrics, logs, and traces. These challenges are particularly pronounced in deep GNNs, where effective multi-hop neighbor aggregation without feature degradation remains difficult.

\subsection{LLM-Based RCA Methods}
Large Language models (LLMs), particularly those based on the Transformer architecture \cite{achiam2023gpt, bai2023qwen, team2023gemini, liu2024deepseek}, have achieved notable success in natural language processing and have recently been extended to the AIOps domain \cite{wu2023large}. Their primary strengths lie in contextual understanding, pattern recognition, and sequence modeling capabilities.

LLMs have been applied to the analysis of unstructured log data for tasks such as log anomaly detection \cite{zhang2025llm}, event clustering \cite{zhu2025cluster}, and failure summarization \cite{kumar2024llms}. By tokenizing data sequences and processing them with LLMs, normal log or metric patterns can be learned, enabling the identification of anomalous patterns. Compared with traditional template-matching or statistical methods, these approaches are more robust and flexible.

More recent studies have explored the reasoning capabilities of LLMs for fault location and question answering. \citet{han2024potential} proposed a hybrid RCA framework that combines a lightweight classifier with an LLM, where the classifier performs preliminary fault identification and is iteratively refined using LLM-generated feedback. \citet{sarda2024augmenting} introduced an LLM-driven framework for cloud-native systems that integrates logs, metrics, traces, and alerts, where alert information is incorporated to enhance contextual learning for automatic identification of incident root cause categories. In addition, the emergence of retrieval-augmented generation (RAG) \cite{gao2023retrieval} and agent-based frameworks \cite{wang2024survey} has further accelerated the application of LLMs in intelligent operations. \citet{ren2025multi} proposed an LLM-based multi-agent system for root cause reasoning that incorporates Monte Carlo Tree Search and a knowledge-based reward mechanism. \citet{roy2024exploring} conducted the first empirical evaluation of an LLM-based ReAct agent for root cause analysis in cloud incident management, demonstrating that comparable performance to RAG and chain-of-thought approaches can be achieved under zero-shot and out-of-domain settings, while fewer factual errors are produced.

Despite their promise, such methods primarily focus on high-level reasoning and summary generation, and rely on intelligent agent tools or knowledge bases, often lacking a comprehensive analysis and understanding of underlying telemetry data. Moreover, hallucination phenomena and error propagation in large language models frequently lead to inaccurate root cause localization results \cite{cornacchia2025between}.

Motivated by these limitations and inspired by ResNet’s ability to fuse multi-level representations through residual connections \cite{he2016deep}, this work investigates the use of residual-connection-enhanced architectures for integrating heterogeneous telemetry data and supporting more fine-grained root cause localization in microservice systems.

\section{Method}
The proposed RC-LLM method for RCA consists of five layers: data input, data preprocessing, data analysis, data integration and LLM reasoning, as illustrated in Fig. 2.

\begin{figure}
    \centering
    \includegraphics[width=1\linewidth]{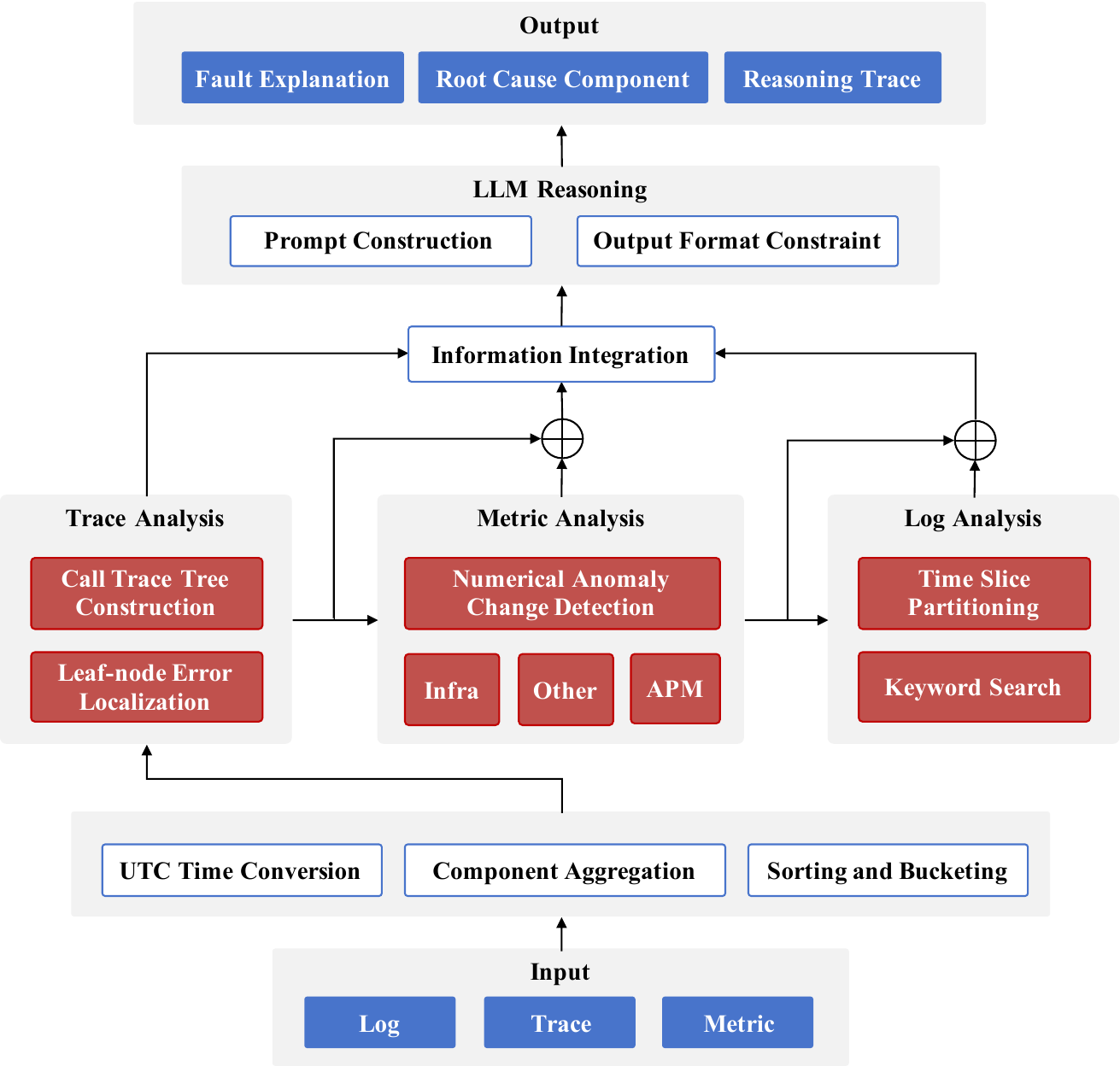}
    \caption{Architecture of RC-LLM.}
    \label{fig2}
    \vspace{-0.5em}
\end{figure}

First, at the data input layer, log, trace, and metric data are loaded from Parquet files using PyArrow. At the data preprocessing layer, all data are converted to unified UTC timestamps, aggregated by component, bucketed at an hourly granularity, and temporally ordered. Next, at the data analysis layer, customized modules perform anomaly detection and feature extraction on metrics, traces, and logs, enabling multi-dimensional data alignment and noise reduction. The resulting structured anomaly information is then provided to the LLM at the reasoning layer, where prompt construction and format constraints are employed for root cause reasoning. Finally, the model outputs are parsed into structured JSON results, including the identified root cause component, failure description, and reasoning path, thereby completing the end-to-end root cause localization process. The details of each layer are described in the following sections.

\subsection{Data Input}
As Parquet is a column-oriented binary storage format with efficient compression and encoding, it enables high-performance reading and processing of large-scale data. Therefore, PyArrow is employed at this layer to efficiently ingest operational telemetry data. Specifically, Parquet files containing log, trace, and metric data are loaded using PyArrow, which fully exploits optimized columnar data access. This data input step can be abstractly formulated as follows:

\begin{equation}
\mathcal{D}=\left\{\mathcal{D}^{trace}, \mathcal{D}^{metric}, \mathcal{D}^{log}\right\}=Load_{PA}\left(P_{s}\right)
\label{eq}
\end{equation}

In the equation (1), $\mathcal{D}$ denotes the complete telemetry dataset, while $\mathcal{D}^{trace}$, $\mathcal{D}^{metric}$, and $\mathcal{D}^{log}$ represents the trace, metric, and log subsets, respectively. $P_{S}$ denotes the Parquet files of each telemetry type and $Load_{PA}(\cdot)$ represents the PyArrow-based loading operation.

This step ensures high data throughput and stable input performance when handling massive monitoring datasets, thereby providing a reliable data foundation for subsequent preprocessing and analysis.

\subsection{Data Preprocessing}
The data preprocessing stage aims to standardize and clean multi-source telemetry data to support subsequent analysis and model reasoning. Log, trace, and metric data are first converted to unified UTC timestamps using NumPy and Pandas, thereby ensuring temporal consistency and eliminating time-zone discrepancies.

Subsequently, data are aggregated and bucketed according to data type and system component. Specifically, telemetry data originating from the same service or pod are grouped together, and different monitoring metrics under the same component, such as CPU and memory usage, are further bucketed. After aggregation and bucketing, all data are temporally ordered to form aligned time series. The data preprocessing procedure can be abstractly formulated as follow:

\begin{equation}
\mathcal{D}_{utc}= \mathcal{F}_{utc\mbox{-}conversion}(\mathcal{D})
\label{eq}
\end{equation}

\begin{equation}
\mathcal{D}_{agg}= \mathcal{F}_{comp\mbox{-}aggregation}(\mathcal{D}_{utc})
\label{eq}
\end{equation}

\begin{equation}
\mathcal{D}^{'}= \mathcal{F}_{sort\mbox{-}bucket}(\mathcal{D}_{agg})
\label{eq}
\end{equation}

In the above equations, $\mathcal{F}_{utc\mbox{-}conversion}$, $\mathcal{F}_{comp\mbox{-}aggregation}$ and $\mathcal{F}_{sort\mbox{-}bucket}$ denote the operations of UTC time conversion, component aggregation, sorting and bucketing, respectively. $\mathcal{D}^{'}$ represent the final preprocessed, temporally aligned time-series data. These steps enable standardized, structured, and temporally coherent data representations, providing high-quality inputs for anomaly extraction and LLM reasoning.

\subsection{Data Analysis}

The data analysis layer serves as the core stage for data convergence and anomaly extraction. Fault-relevant information is identified from large-scale preprocessed data to suppress noise and improve the efficiency and accuracy of large language model reasoning, while reducing context token consumption. At this stage, customized analysis tools are applied to trace, log, and metric data.

\subsubsection{Trace Analysis}
In the trace analysis stage, an iterative hierarchical construction algorithm is employed to build complete service–pod call trees using the Anytree library, as shown in Table 1. By traversing the trees and analyzing gRPC status codes, anomalous leaf nodes and critical invocation segments are efficiently identified and extracted. The trace analysis process can be abstracted as follows:

\begin{equation}
\mathcal{A}_{trace}= \mathcal{F}_{trace}(\mathcal{D}^{'}_{trace})
\label{eq}
\end{equation}

In the equation (5), $\mathcal{D}^{'}_{trace}$ denotes the preprocessed trace data, and $\mathcal{F}_{trace}(\cdot)$ represents the hierarchical trace analysis function. Table I presents an example of a call tree constructed using the Anytree library, in which bold italic pods denote leaf nodes. The call tree explicitly represents inter-service invocation relationships and supports precise identification of faulty services and invocations, thereby providing reliable cues for subsequent root cause localization.

\begin{table}[t]
\small
\renewcommand\arraystretch{1.5}
\tabcolsep=0.3cm
\caption{Example of a Call Chain Tree.}
\label{tab1}
\begin{tabular}{p{8cm}}
\hline
Call Chain Tree Constructed Using Anytree  \\ \hline
\begin{forest}
  for tree={
    font=\ttfamily,
    grow'=0,
    folder,
    l sep=30pt,      % 节点到线的距离
    l=3em,         % 这就是你要的“横线长度”！数值越大，横线越长
    s sep=6pt,
    edge={darkgray, line width=0.5pt}, % 还可以控制线的粗细和颜色
  }
  [frontend-0
    [frontend-0]
    [recommendationservice-1
    [productcatalogservice-1]]
    [cartservice-0
    [\textbf{\emph{redis-cart-0}} \quad \quad \textbf{\emph{\#Faulty Node}}]]
  ]
\end{forest}
  \\ \hline
\end{tabular}
\vspace{-1.0em}
\end{table}

\subsubsection{Metric Analysis}
The metric analysis part focuses on identifying anomalous indicators associated with fault time windows. A combination of statistical thresholding and trend analysis is applied to detect service-level anomalies. Key performance metrics, including CPU utilization, memory usage, request-response time (RRT), request-response pairing deviations, client error rate, server error rate, and timeout counts, are analyzed to capture abnormal service behavior. In addition, the PELT (Pruned Exact Linear Time) algorithm for change point detection is applied to these metrics to identify abrupt numerical changes across consecutive time intervals. The metric analysis process can be abstracted as follows:

\begin{equation}
\mathcal{A}_{metric}= \mathcal{F}_{metric}(\mathcal{D}^{'}_{metric})
\label{eq}
\end{equation}

In the above equation, $\mathcal{D}^{'}_{metric}$ denotes the preprocessed trace data, and $\mathcal{F}_{metric}(\cdot)$ denotes the metric analysis function that detects abnormal indicators using statistical thresholding, trend analysis, and change point detection.

\begin{figure}[t]
    \centering
    \includegraphics[width=1\linewidth]{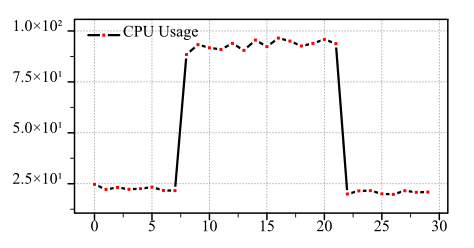}
    \caption{Sudden Numerical Anomaly of  CPU Usage.}
    \label{fig3}
    \vspace{-0.5em}
\end{figure}

\begin{figure}[t]
    \centering
    \includegraphics[width=1\linewidth]{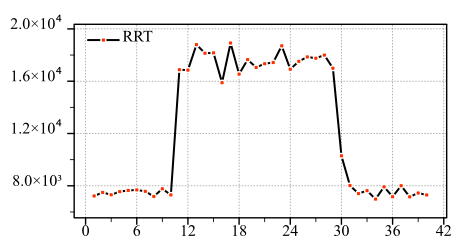}
    \caption{Sudden Numerical Anomaly of RRT.}
    \label{fig4}
    \vspace{-0.5em}
\end{figure}

\begin{figure}[t]
    \centering
    \includegraphics[width=1\linewidth]{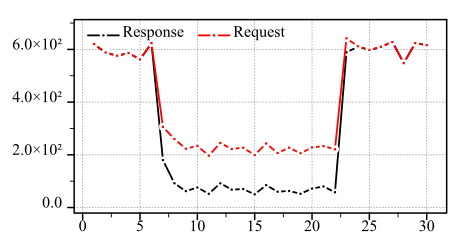}
    \caption{Request-response mismatch Anomaly.}
    \label{fig5}
    \vspace{-0.5em}
\end{figure}

Figs. 3 to 5 present representative examples of abrupt numerical anomalies observed in different performance metrics, including CPU usage stress anomalies, sudden spikes in request-response time (RRT), and mismatches between request and response volumes. These anomalies reflect distinct types of abnormal behavior during fault periods. When significant fluctuations or outliers are detected, the corresponding components and their associated metrics are identified and extracted, enabling the selection of services or pod nodes with potential performance issues and providing quantitative evidence for large language model reasoning.

\subsubsection{Log Analysis}
In the log analysis stage, anomalous log entries are first extracted from log data. Predefined pattern-matching keywords are applied to filter log records containing error-related information. Representative keywords include Error, Timeout, Exception, Failed, and common HTTP status codes (e.g., 400, 404, and 500). Log anomaly extraction is abstracted as a fault-related information filtering process.

\begin{equation}
\mathcal{A}_{log}= \mathcal{F}_{log}(\mathcal{D}^{'}_{log})
\label{eq}
\end{equation}

In the equation (7), $\mathcal{D}^{'}_{log}$ denotes the preprocessed log data, and $\mathcal{F}_{log}(\cdot)$ filters fault-related log entries based on predefined error patterns and keywords. This filtering process substantially reduces the volume of log data, enabling the LLM to focus on fault-relevant information rather than redundant log entries.

\begin{table*}[h]
\small
\centering
\renewcommand\arraystretch{1.7}
\tabcolsep=0.3cm
\caption{Statistics of dataset used in our experiments.}
\label{tab2}
\begin{tabular}{c|c|c}
\hline
\textbf{Data}   & \textbf{Included Information} & \textbf{Records} \\ \hline
Trace  & spanID, traceID, references, process, duration, tags &  121,996,403   \\ \hline
Metric &   CPU usage, memory usage, disk read bytes, network transmit, request, rrt, timeout &  18,815,081  \\ \hline
Log    &  message, detailed textual information generated during system operation   &   128,481,689     \\ \hline
\end{tabular}
\vspace{-1.0em}
\end{table*}

\subsection{Data Integration}
After multi-dimensional anomalies are identified at the data analysis layer, the extracted data are not directly concatenated. Instead, a hierarchical integration mechanism inspired by residual connections is adopted to fuse multi-source information and construct a compact yet highly informative evidence context for LLM reasoning. The overall workflow is illustrated in Fig. 6.

\begin{figure}[h]
    \centering
    \includegraphics[width=1\linewidth]{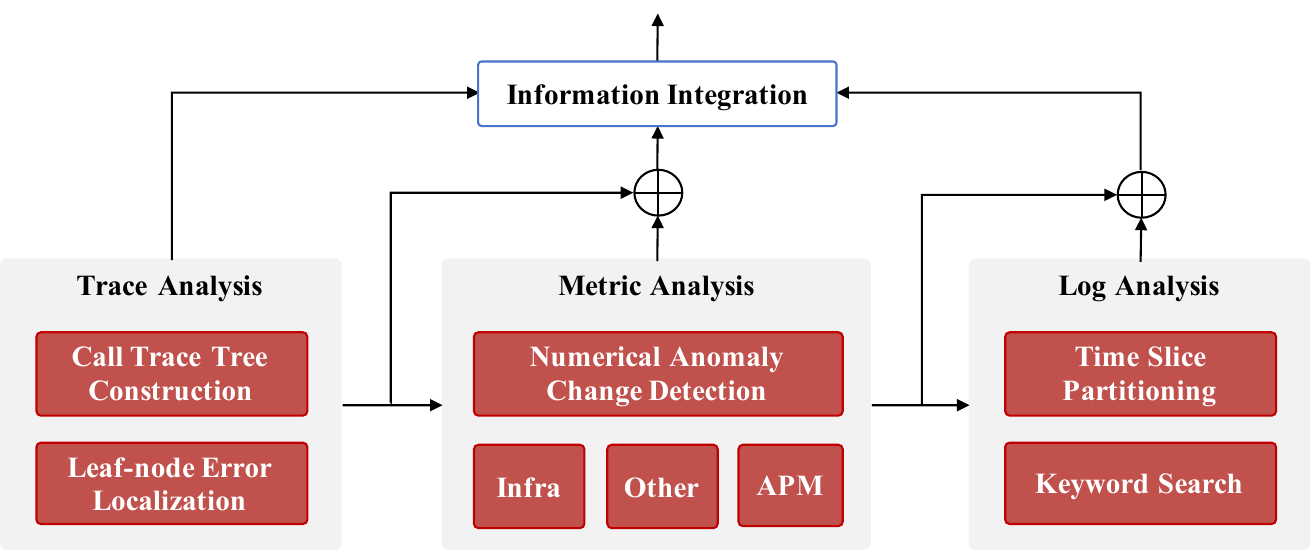}
    \caption{Data Integration Based on a Residual Structure.}
    \label{fig6}
\end{figure}

The integration process is initiated with trace data as the primary anchor. When an erroneous component is detected in the call chain, it is selected as the focal point for subsequent analysis.

Next, metric data corresponding to the identified component are examined with priority. Key indicators, such as request-response time (RRT), error rate, and CPU utilization, are evaluated within the fault time window to determine whether abnormal fluctuations are also observed. This step enables quantitative cross-validation of trace-level anomalies and provides multi-dimensional evidence.

Subsequently, log data generated by the same component are inspected to extract anomalous entries, including error logs, exception stacks, or specific failure patterns. By associating these logs with trace and metric information, detailed textual descriptions of the failure are provided, facilitating a deeper understanding of the fault context.

Finally, when anomalies related to the same component are available across trace, metric, and log data, they are jointly integrated to form a complete and highly reliable evidence chain for LLM inference. In cases where certain data modalities are missing (e.g., unavailable metric data), all accessible anomalous information is flexibly integrated. Whether composed of multi-source or single-source data, the resulting data packages are supplied to the LLM for unified reasoning, ensuring effective root cause localization under varying data completeness conditions. The data integration process can be abstractly formulated as follows:

\begin{equation}
E_{integrated}= \mathcal{A}_{trace} \oplus \mathcal{A}_{metric} \oplus \mathcal{A}_{log}
\label{eq}
\end{equation}

In the equation (8), $E_{integrated}$ denotes the integrated context data. $\oplus$ denotes a hierarchical residual-based fusion operation that preserves dominant fault signals while allowing flexible incorporation of available modalities. Through this strategy-driven integration of trace, metric, and log analysis results, fault-relevant information is effectively distilled from large-scale monitoring data. Noise is substantially reduced, and high-quality inputs are provided for LLM inference, thereby improving both the efficiency and accuracy of root cause localization.

\subsection{LLM Reasoning}

To ensure interpretability and reliability of the diagnostic results, structured output constraints are enforced using the \emph{PromptTemplate} and \emph{StructuredOutputParser} tools from LangChain \cite{mavroudis2024langchain}. These tools constrain the JSON format and reasoning structure of both LLM inputs and outputs. The LLM is required to produce outputs containing three predefined fields: the root cause component, the fault explanation, and the reasoning trace.  The constrained LLM reasoning and output generation process can be abstractly expressed as follows:

\begin{equation}
O=\left\{C_{r}, F_{e}, R_{t}\right\}=LLM(E_{integrated})
\label{eq}
\end{equation}

In the equation (9), $O$ denotes the structured output generated by the LLM, $C_{r}$ denotes the identified root cause component, $F_{e}$ denotes the fault explanation in natural language, and $R_{t}$ denotes the reasoning trace that explicitly records the inference process from observed anomalies to the final conclusion, thereby improving result interpretability and traceability. These structured outputs are directly used as the final results for root cause localization.

\section{Experiment}
This chapter aims to comprehensively validate the proposed approach through empirical evaluation and case studies. First, the datasets used in the experiments are introduced. Next, the iterative optimization process of the proposed method is presented to demonstrate progressive improvements in performance. Finally, representative case studies are analyzed to illustrate the effectiveness of the approach in identifying root causes from multi-dimensional telemetry data.

\subsection{Dataset}
The experimental evaluation is conducted on the public dataset released for the CCF AIOPS 2025 Root Cause Analysis Challenge. The dataset is constructed and extended based on HipsterShop, an open-source microservice system provided by Google, and is designed to simulate the dynamics and complexity of real-world microservice systems, thereby providing a representative experimental environment for root cause analysis. The statistics of trace, metric, and log data in the dataset are summarized in Table II.

The microservice system adopts a dynamically deployed architecture consisting of 10 core business microservices and 8 virtual machines (VMs). Each core microservice is deployed with three Pods, resulting in a total of 30 Pods that are dynamically scheduled across the 8 VMs by Kubernetes. In addition, a TiDB database cluster is integrated into the system, including the core services tidb-tidb, tidb-pd, and tidb-tikv, each deployed with a single Pod on the VMs. To construct diverse and realistic failure scenarios, the dataset incorporates a fine-grained fault injection mechanism that supports three levels of faults: service-level, pod-level, and node-level. The dataset contains 400 fault cases, in which the model is required to identify the corresponding root cause within a given time window. 

It is worth noting that the RCA task in this study differs from conventional operational RCA settings. In practical monitoring systems, RCA is typically triggered by explicit alarms that provide an initial anchor for fault tracing, whereas the dataset used in this work does not include predefined alarm locations or hints. Consequently, potential faulty components must be inferred solely from observed telemetry characteristics and heuristic rules, without relying on prior alarm anchors. This setting substantially increases the difficulty of root cause localization, as the method is required to first identify components involved in fault propagation before reasoning about the root cause.

\begin{table*}[h]
\small
\centering
\renewcommand\arraystretch{1.7}
\tabcolsep=0.3cm
\caption{Iterative Evolution of the Method.}
\label{tab3}
\begin{tabular}{>{\centering\arraybackslash}m{2.5cm}>{\centering\arraybackslash}m{10cm}>{\centering\arraybackslash}m{1.5cm}>{\centering\arraybackslash}m{1.5cm}}
\hline
\textbf{Strategy}   & \textbf{Description} & \textbf{Accuracy(\%)} & \textbf{Avg Steps} \\ \hline

Original  & Log, Metric, and Trace data are directly fed into the LLM. &  23.75 & 6.2  \\ \hline

Early-stage & Abnormal APM components are extracted, and their corresponding Trace and Log data are fed into the LLM. &  28.50 & 4.7 \\ \hline

Intermediate-stage &  Abnormal gRPC components are extracted from the Trace data, and their corresponding APM and Log data are fed into the LLM.   &   41.75  & 4.3     \\ \hline

\textbf{Final (RC-LLM)} &  From Trace to Metric to Log, residual fusion is performed at each level, and the fused representations are fed into the LLM.   &   \textbf{48.25}  & \textbf{3.4}   \\ \hline

\end{tabular}
\vspace{-1.0em}
\end{table*}

\subsection{Experimental Details}

The methods are evaluated using Accuracy and average reasoning steps to assess the effectiveness and efficiency of root cause localization, respectively. Accuracy measures the correctness of identifying the true root cause component, while the average number of reasoning steps reflects the reasoning efficiency of the methods. 

Although RAG and external knowledge bases are known to improve performance in many tasks, they introduce additional dependencies on the completeness and quality of the constructed knowledge base. To better highlight the inherent capability and architectural advantages of the methods, no retrieval enhancement based on external knowledge bases is incorporated in the experiments.

Furthermore, after a comprehensive evaluation of token consumption, inference quality, and response latency, \textbf{\emph{DeepSeek-V3 (671B)}} \cite{liu2024deepseek}was selected as the core reasoning engine. 

For experimental fairness and consistency, all reasoning relies solely on observed telemetry data without incorporating external knowledge retrieval, and the same DeepSeek-V3 (671B) model is used throughout all experiments. The implementation is developed in Python and executed in a Linux-based environment.

\subsection{Method Iteration and Analysis}

The RC-LLM framework was developed through four successive iterations. The key modifications and accuracy of each iteration are summarized in Table III and described below.

\subsubsection{Original}
Log, Trace, and Metric data within the fault time window were minimally processed and directly fed into the LLM for reasoning. This naive strategy achieved an accuracy of 23.75\%. Experimental results revealed clear limitations of this naive approach, including context-length overflow, low accuracy, and high response latency.

\subsubsection{Early-stage}
Abnormal service or pod components were first identified based on anomalous APM metrics within the fault window, and the corresponding trace and log data were subsequently supplied to the LLM for reasoning. This strategy improved the localization accuracy to 28.50\%, showing a moderate gain over the original approach. However, the results indicated that components exhibiting metric anomalies were not necessarily the root causes, highlighting the need for additional evidence sources to improve localization accuracy.

\subsubsection{Intermediate-stage}
The metric-driven component selection was abandoned, and trace call chains were instead adopted as the primary analysis entry point. Nodes with abnormal gRPC status codes were identified, and candidate root components, as well as fault propagation paths, were inferred based on parent-child relationships. Associated APM metrics and log data were then incorporated for LLM reasoning. This strategy further increased the accuracy to 41.75\%, demonstrating the effectiveness of trace-based analysis for root cause localization. Nevertheless, certain failure cases were observed without abnormal gRPC status codes, indicating incomplete coverage of this strategy.

\subsubsection{Final}
To address failures with normal gRPC status codes, the intermediate strategy was further enhanced by integrating metric-based anomaly detection. Service or pod components exhibiting significant metric anomalies or abrupt metric changes were selected from APM data and residually fused with log entries containing error-related keywords (e.g., Error, Exception). The fused multi-source information was then provided to the LLM for reasoning, enabling effective diagnosis in scenarios where trace-based signals were insufficient. This final strategy achieved an accuracy of 48.25\%, demonstrating a substantial improvement over all previous stages and validating the effectiveness of the proposed RC-LLM framework.

\subsection{Case Study}
To further evaluate the proposed method, a representative case from the dataset is analyzed to illustrate the diagnostic process and reasoning behavior of the model. The detailed analysis results are presented in Table IV.

\begin{table*}[t]
\centering
\small
\renewcommand\arraystretch{1.5}
\tabcolsep=0.3cm
\caption{A Case Study of Root Cause Analysis.}
\label{tab4}
\begin{tabular}{>{\centering\arraybackslash}m{2.5cm}|>{\arraybackslash}m{14cm}}
\hline
 \textbf{Question} & A fault occurred from 2025-06-05T18:10:05Z to 2025-06-05T18:34:05Z. Please identify the root cause.
\\ \hline
\textbf{Ground Truth} & component: cartservice \newline
fault description: ["pods crash", "pods terminated unexpectedly", "container exit", "pods failure"] \newline
key metrics:["client error","client error ratio","error","error ratiopod processes", "request", "response","RRT"]
\\ \hline
\textbf{Ours} & component: cartservice \newline
reason: response timeout with RRT spike to 97k ms \newline 
reasoning trace:[ \newline
{"step":1,"action": "TraceAnalysis(Tracedata)", "observation": "3 cartservice pods showing errors"} \newline 
{"step":2,"action":"MetricsAnalysis(cartservice)","observation":"RRT spike to 97246ms at fault time"} \newline
{"step":3,"action": "LogSearch(cartservice)", "observation": "40 errors detected at faultpeak"} \newline
{"step":4,"action": "AnalyzeAPM(cartservice)", "observation": "23.12\% error ratio at 18:10"}] \\ \hline
\end{tabular}
\vspace{-1.0em}
\end{table*}

As shown in Table IV, the proposed method correctly localizes the root cause component, demonstrating the effective coordination of the designed techniques. In the TraceAnalysis step of the reasoning trace, three anomalous cartservice pods are first identified based on abnormal gRPC status codes, enabling accurate localization of faulty leaf nodes in the call chain. Subsequently, in the MetricAnalysis step, a significant spike in request-response time (RRT) of 97246 ms is detected for cartservice, as illustrated in Fig. 7, further corroborating the presence of service-level performance degradation. In the LogSearch step, 40 error-related log entries are extracted, and abnormal error ratios in APM metrics are additionally confirmed. All observations are consistent with the ground truth.

\begin{figure}[t]
    \centering
    \includegraphics[width=1\linewidth]{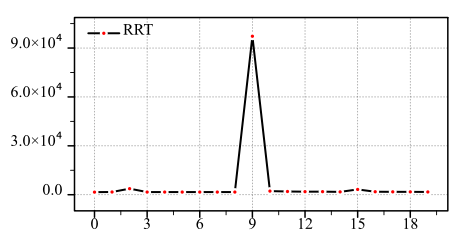}
    \caption{A Sudden Numerical Anomaly of  RRT.}
    \label{fig7}
    \vspace{-0.5em}
\end{figure}

These observations are integrated into a coherent reasoning chain—pod errors → RRT spike → error logs—which supports reliable root cause localization. The final output correctly reports the root cause component in the predefined JSON format. It is worth noting that each service in the system is composed of three pods. In this case, anomalies are simultaneously detected across all three cartservice pods, indicating that the failure is not an isolated pod-level issue but a service-level fault. The proposed method captures this multi-pod anomaly pattern and leverages inter-pod correlation to infer the service-level nature of the failure, thereby enabling accurate root cause identification.

In some cases, the proposed method exhibits reduced accuracy at the node-level root cause localization. Although service-level and pod-level faults can be reliably identified through the integration of Trace and Metric information, insufficient fine-grained metric discrimination and limited log-level semantic evidence hinder precise differentiation among multiple replicas of the same service, resulting in coarse-grained reasoning and node-level misidentification.

\section{Conclusion}
In this paper, a residual-connection-based LLM approach, termed RC-LLM, is proposed for root cause analysis in large-scale microservice systems. Root cause analysis is reformulated as a temporal causal reasoning problem, in which Trace, Metric, and Log data are hierarchically integrated through a residual fusion mechanism. This design preserves critical fault signals while suppressing noise introduced by fault propagation. Experimental results on the CCF AIOps dataset show that RC-LLM achieves strong performance in terms of root cause localization accuracy and inference efficiency. Moreover, case studies indicate that the proposed approach is capable of constructing coherent reasoning chains and delivering interpretable diagnostic results across diverse failure scenarios.

\bibliographystyle{IEEEtranN}
\bibliography{Conference-LaTeX-template_10-17-19/mybib}

\end{document}